\documentclass{article}
\usepackage[preprint]{colm2026_conference}

\usepackage{microtype}
\usepackage{hyperref}
\usepackage{url}
\usepackage{booktabs}
\usepackage{enumitem}
\usepackage{tikz}
\usepackage{graphicx}
\usepackage{caption}
\usepackage{subcaption}
\usepackage{float}
\usepackage{amsmath}
\usetikzlibrary{positioning, arrows.meta, shapes.geometric, fit, backgrounds, decorations.pathreplacing, patterns}
\usepackage{placeins}
\usepackage{lineno}

\definecolor{darkblue}{rgb}{0, 0, 0.5}
\hypersetup{
  colorlinks=true,
  citecolor=darkblue,
  linkcolor=darkblue,
  urlcolor=darkblue
}

\title{Evaluating Memory Condensation Strategies for Coding Agents in Data-Driven Scientific Discovery}

\author{%
\begin{minipage}{\textwidth}\centering
{\bf Renuka Chintalapati}$^{1}$, {\bf Sid Raskar}$^{1}$, {\bf Anurag Acharya}$^{1}$, \\[2pt]
{\bf Jared Willard}$^{2}$, {\bf Patrick Emami}$^{2}$, {\bf Sameera Horawalavithana}$^{1}$ \\[4pt]
$^{1}$Pacific Northwest National Laboratory \quad $^{2}$National Laboratory of the Rockies \\[4pt]
{\normalfont\texttt{renuka.chintalapati@pnnl.gov}, \texttt{s.raskar@pnnl.gov}, \texttt{anurag.acharya@pnnl.gov},} \\
{\normalfont\texttt{jared.willard@nlr.gov}, \texttt{patrick.emami@nlr.gov}, \texttt{yasanka.horawalavithana@pnnl.gov}}
\end{minipage}%
}

\begin{document}

\ifcolmsubmission
\linenumbers
\fi

\maketitle

\begin{abstract}

Coding agents accumulate extensive context during long-running tasks, yet fixed context windows force practitioners to choose between truncation and task failure. While numerous memory condensation strategies have been proposed from simple sliding windows to LLM-generated summaries, no systematic comparison exists to guide strategy selection, especially in scientific discovery tasks. We evaluate 8 memory condensation strategies using GPT-4o on 60 DiscoveryBench tasks spanning 6 scientific domains (480 total evaluations). We find that no condenser significantly alters hypothesis quality, while LLM‑based condensers increase token costs by 24–94\%, and masking the tool‑call outputs achieves an 8.6\% net savings. 
We also noticed that the optimal condenser applied in data-driven scientific discovery tasks varies by scientific domain and task length.

\end{abstract}

\section{Introduction}
  \label{sec:introduction}

As large language models (LLMs) transition from single-turn tools to autonomous agents that pursue goals across
extended interactions~\citep{Sumers2024cognitive}, context becomes central to reliable reasoning. Relevant information
keeps agents aligned with their goals, while missing or incomplete context leads to drift or hallucination. This has
elevated ``context engineering''~\citep{Mei2025survey} into a growing discipline focused on managing long interaction
histories. However, agents accumulate context rapidly from tool outputs to files and error logs, often reaching
tens or hundreds of thousands of tokens~\citep{Yao2022react}. Even with larger context windows, models still degrade on
long inputs~\citep{liu2024lost}, and costs scale with context length. Users must therefore choose between
retaining all information at high computational cost, or compressing aggressively and risk losing important
details.

Memory condensation strategies offer a practical alternative to naive truncation. These methods selectively compress
or preserve important context: sliding-window techniques keep recent events, masking strategies hide verbose outputs,
and summarization methods distill long histories into compact representations~\citep{Zhang2024survey,Du2025rethinking,Packer2024memgpt}.  
Sliding windows prioritize recency, summarization targets semantic
compression~\citep{Liu2023thinkmemory}, and attention-based approaches rely on the model's ability to identify salient
content. This raises a key question:~\textit{which of these memory condensation approaches leads to better performance?}

Despite growing interest, systematic comparison of condensation strategies remains limited. Recent benchmarks evaluate
agent memory quality~\citep{Bian2026realmem,Chen2025halumem,Deshpande2025memtrack,Tang2026memoryrewardbench},
cost-aware planning~\citep{Liu2025costbench,Wolff2026costaccuracy}, framework-level
efficiency~\citep{Yin2025agentframeworks,Roig2025enterprise}, prompt caching~\citep{Lumer2026promptcache}, and
performance under growing context~\citep{Fang2026agentlongbench}, while~\cite{Dong2025compressedagents} find that
compressed LLMs degrade on complex tasks. However, none of these works compare multiple condensation strategies while
controlling for task type and model. Scientific discovery is a particularly compelling setting for this comparison:
agents must maintain coherent, multi-step reasoning chains, remembering earlier statistical findings, variable
relationships, and failed hypotheses across long data-exploration sessions. 

To this end, we evaluate eight memory condensation strategies on DiscoveryBench~\citep{Majumder2024discoverybench}, a benchmark focused on data-driven discovery tasks spanning six scientific
domains---Biology, Economics, Engineering, Humanities, Meta Science, and Sociology. 
These tasks mirror real
scientific-coding workloads, requiring both software engineering (e.g., data processing, debugging) and scientific
reasoning (e.g., hypothesis formation, statistical interpretation). 

Our evaluation addresses three research questions:
\textbf{RQ1:} \textit{How does memory condensation impact task completion rates, token savings and solution quality?}
\textbf{RQ2:} \textit{Do different scientific domains benefit from different condensation approaches?}
\textbf{RQ3:} \textit{How do token costs scale with task length across condensation strategies?}

\textbf{Our contributions:}
\textbf{(1)} We conduct a systematic evaluation of 8 condensation strategies (5 heuristic, 3 LLM-based) across 480
tasks in 6 scientific domains;
\textbf{(2)} We show that masking the tool-call outputs outside a recent attention window
with compact placeholders records the highest efficiency, achieving 8.6\% token savings without LLM overhead;
\textbf{(3)} We find that no condenser produces statistically significant changes in hypothesis quality while LLM-based condensers increase total token usage by 24--94\% due to condensation overhead and additional agent turns meaning they incur substantial cost without measurable quality benefit. 
\textbf{(4)} We demonstrate strong domain dependence where the tool-call output masking method that saves up to 52\% tokens in verbose domains
like sociology, only provides limited benefit in concise Meta Science tasks.
On the other hand, LLM-based condensers record greater
quality gains in harder domains (e.g., biology, economics) and heuristic methods suffice in simpler ones.

\section{Agentic Memory Condensation Techniques}
\label{sec:background}


In this study, we used OpenHands~\citep{OpenHands2024} (formerly OpenDevin), an open-source platform for LLM-powered software development agents, enabling code execution, file operations, web browsing, and human communication. 
We selected OpenHands because its open-source nature provides transparency into agent internals for reproducible experimentation, and its modular \textit{Condenser} interface allows systematic comparison of condensation strategies without modifying core agent logic.


As agents engage in long-running tasks, conversation histories grow substantially, leading to increased API costs and degraded effectiveness~\citep{Mei2025survey}. OpenHands represents these histories as a stream of \emph{events}, discrete units comprising agent actions (code execution, file operations, commands) and observations (tool outputs, error messages, results). OpenHands addresses context growth through a condenser abstraction that maintains bounded context while preserving essential information~\citep{Packer2024memgpt,Chhikara2025mem0}. The framework implements multiple strategies varying along two dimensions: \emph{compression mechanism} (heuristic vs. LLM-based) and \emph{computational overhead} (no additional LLM calls vs. requiring inference)~\citep{Du2025rethinking,Shan2025cognitive}. 

Condensers are configured via parameters: \textit{keep\_first} specifies how many initial events (typically system prompt and task description) to always preserve; \textit{max\_size} sets the event count threshold that triggers condensation; \textit{max\_events} bounds the total events retained; and \textit{attention\_window} controls how many recent observations remain unmasked.

\textbf{Heuristic Strategies} apply deterministic rules without additional LLM inference~\citep{Du2025rethinking}. The simplest is \textit{NoOp} condenser, a pass-through baseline that preserves all events and serves as our upper bound for information retention. Two strategies implement variations of sliding window eviction: \textit{RecentEvents} condenser maintains a fixed-size FIFO window retaining \textit{keep\_first} initial events plus the most recent events up to \textit{max\_events} total, while \textit{AmortizedForgetting} condenser triggers when count exceeds \textit{max\_size} and aggressively reduces to \textit{max\_size/2} events, preserving head (task setup) and minimal tail (recent context). \textit{ObservationMasking} condenser targets verbose observations rather than event count: it retains all events in the history
but replaces tool output content outside an \textit{attention\_window} with placeholder tokens, preserving the full event sequence while reducing token volume. Finally, \textit{ConversationWindow} condenser preserves essential dialogue
structure and triggers reactively on overflow errors rather than proactively on size thresholds.

\textbf{LLM-Based Strategies} invoke a secondary model for intelligent compression~\citep{Zhong2023memorybank,Liu2023thinkmemory}, trading computational overhead for semantic understanding.
\textit{LLMSummarizing} condenser generates free-form summaries that capture user goals, completed and pending tasks, code state, and error history---replacing dropped events with a condensed narrative. 
\textit{StructuredSummary} condenser produces structured summaries with 18 predefined fields (task status, file modifications, test results, git state) via function calling, ensuring consistent information extraction. 
\textit{LLMAttention} condenser takes a different approach: rather than summarizing, it queries the model to rank all events by relevance to the current task, retaining the highest-ranked events regardless of recency~\citep{Yan2025memoryr1}. 
These LLM strategies use GPT-5-mini for condensation.

Condensers follow three triggering modes: \emph{proactive} methods (\textit{RecentEvents, Amortized}, and all LLM‑based) run once the event count exceeds a threshold; \emph{per‑turn} methods (\textit{ObservationMasking}) apply transformations every iteration; and \emph{reactive} methods (\textit{ConversationWindow}) trigger only when a context‑overflow error occurs.

\section{Experimental Setup}
\label{sec:setup}

We design experiments to evaluate how condensation strategies affect agent performance on scientific reasoning tasks. Our evaluation framework measures computational efficiency, enabling comparison of the 8 condensation strategies described in Section~\ref{sec:background}.

\subsection{Implementation}
\label{sec:implementation}

\paragraph{Agent Framework}                                                                                     
We use the CodeActAgent~\cite{wang2024executable}, OpenHands' default agent architecture that interleaves reasoning with executable code actions and no modifications were made to the agent itself. 

\paragraph{Runtime Environment}
Agents execute in isolated Docker containers using the OpenHands runtime image (\texttt{oh\_v1.2.1}) with \texttt{python:3.12-bookworm} as the base image. Each task runs in a fresh container with standardized dependencies including pandas, numpy, scipy, scikit-learn, and statsmodels for data analysis tasks. The sandboxed environment ensures reproducibility and prevents cross-task contamination.

\paragraph{LLM Integration}
We interface with language models through LiteLLM (v1.74.3+), providing unified API access across providers. All agent experiments use GPT-4o via Azure OpenAI Service as the backbone model. Calls use \texttt{temperature=0.0} for deterministic outputs, \texttt{num\_retries=8} with exponential backoff, and \texttt{caching\_prompt=true} where supported. LLM-based condensers use GPT-5-mini for condensation, a smaller and cost efficient model than the main agent. Hypothesis quality evaluation utilizes GPT-4o as the LLM Judge model. 

\subsection{Evaluation Tasks}
\label{sec:tasks}

\paragraph{DiscoveryBench}
We use DiscoveryBench~\citep{Majumder2024discoverybench}, which evaluates LLM capabilities in data-driven scientific discovery. The dataset comprises 144 real-world tasks across six domains---Biology, Economics, Engineering, Humanities, Meta Science, and Sociology---each requiring agents to analyze datasets and generate testable hypotheses derived from published research. Each task provides: (1)~a scientific query requiring data analysis, (2)~one or more datasets (CSV/JSON) with real-world measurements, (3)~metadata describing variables and relationships, (4)~a gold hypothesis for evaluation, and (5)~workflow tags indicating expected analytical approaches.

\begin{table*}
\centering
\begin{tabular}{@{}llrl@{}}
  \toprule
  \textbf{Domain} & \textbf{Tasks} & \textbf{Avg Ev.} & \textbf{Primary Challenge} \\
  \midrule
  Biology & 10 & 11.0 & Statistical tests \\
  Engineering & 10 & 10.7 & Requirements tracing \\
  Sociology & 10 & 10.2 & Causal inference \\
  Humanities & 10 & 8.6 & Archaeological patterns \\
  Economics & 10 & 8.1 & Econometric modeling \\
  Meta Science & 10 & 5.4 & Literature synthesis \\
  \bottomrule
\end{tabular}
\caption{DiscoveryBench domain characteristics. Event counts from NoOp baseline.}
\label{tab:domain-characteristics}
\end{table*}

We use DiscoveryBench because its open-ended scientific reasoning tasks introduce realistic memory-management challenges not present in code-focused benchmarks. Tasks range from 5--22 events, requiring agents to maintain coherent, long-horizon reasoning chains across diverse domains. Data-intensive fields such as Engineering and Sociology produce verbose file interactions whereas compact domains like Meta Science and Humanities require preserving inferential structure within shorter contexts. 

Table~\ref{tab:domain-characteristics} highlights these differences: Biology tasks average 11.0 events with complex statistical analyses; Engineering averages 10.7 events with requirements tracing; and Meta Science tasks, though shortest at 5.4 events, demand precise literature synthesis. Our evaluation uses stratified sampling: 10 tasks per domain (test split only) across six domains (60 total) yielding 480 evaluations in total ($60 \times 8$ condensers).

\subsection{Evaluation Metrics}
\label{sec:metrics}

Our evaluation captures both computational efficiency and hypothesis quality.

\paragraph{Efficiency Metrics}
We measure efficiency along multiple dimensions: \emph{token usage} (total input and output tokens consumed, reported as average per instance and percentage savings relative to \textit{NoOp} baseline), \emph{event count} (number of agent interactions to task completion, with task-level breakdown of faster/same/slower than baseline), and \emph{input/output ratio} (decomposition into prompt vs.\ completion tokens).
We also track \emph{token accumulation} (cumulative growth revealing linear vs.\ quadratic cost scaling), and \emph{domain-specific performance} (per-domain savings to identify which task types benefit from each condenser).

\paragraph{Hypothesis Quality (LLM-as-Judge)}
We assess hypothesis quality using an LLM-as-Judge framework adapted from DiscoveryBench~\citep{Majumder2024discoverybench}. A judge model (GPT-4o via Azure OpenAI) evaluates generated hypotheses against gold standards across three dimensions: \emph{Context Matching} (ternary: 1.0 for very similar, 0.5 for partial match, 0.0 for different temporal/conditional/scope context), \emph{Variable Overlap} (F1 score with fuzzy matching for equivalent variable names), and \emph{Relationship Similarity} (ternary: 1.0 for very similar, 0.5 for more general, 0.0 for different). 
The final score weights context most heavily, as correct scope and conditions are prerequisites for a valid scientific hypothesis: $\text{Score} = 0.4 \times \text{Context} + 0.3 \times \text{Variable\_F1} + 0.3 \times \text{Relation}$.

\paragraph{Few-Shot Judge Calibration}
We augment the relationship similarity and context matching dimensions with few-shot examples that clarify statistical equivalences (e.g., showing that a “positive relationship’’ corresponds to a correlation of +0.66), preventing penalties for scientifically precise hypotheses. Since our primary objective is to compare condensation strategies, we incorporated a robust LLM judge and conducted manual spot checks of its outputs to verify that scoring remained reasonable across condensers.



\subsection{Experimental Protocol}
\label{sec:protocol}

\paragraph{Condenser Configuration}
We use a two-stage calibration process to determine condenser settings. First, we run the \textit{NoOp} (no-condensation) baseline across all 60 tasks to characterize uncondensed behavior, which averages 9 events per task (range: 5--22) and roughly 75k tokens. Based on these statistics, we set \texttt{max\_size=8}, ensuring condensation activates only for tasks exceeding the median event count, thereby avoiding unnecessary compression on shorter tasks while focusing on cases where memory pressure is most significant.

We preserve the first two messages (\texttt{keep\_first=2}) to retain the system prompt and task description, satisfying the constraint \texttt{keep\_first < max\_size/2}. For \textit{ObservationMasking}, we set \texttt{attention\_window=5}, which keeps the five most recent observations unmasked. Table~\ref{tab:condenser-config} summarizes the full configuration.

\begin{table*}
\centering
\begin{tabular}{@{}lll@{}}
  \toprule
  \textbf{Condenser} & \textbf{Type} & \textbf{Parameters} \\
  \midrule
  NoOp & Baseline & --- \\
  RecentEvents & Heuristic & max\_events=8, keep\_first=2 \\
  Amortized & Heuristic & max\_size=8, keep\_first=2 \\
  ObservationMasking & Heuristic & attention\_window=5 \\
  ConversationWindow & Heuristic & (reactive trigger) \\
  \midrule
  LLMSummarizing & LLM-based & max\_size=8, keep\_first=2 \\
  StructuredSummary & LLM-based & max\_size=8, keep\_first=2 \\
  LLMAttention & LLM-based & max\_size=8, keep\_first=2 \\
  \bottomrule
\end{tabular}
\caption{Condenser parameters calibrated from NoOp baseline (avg 9.0 events).}
\label{tab:condenser-config}
\end{table*}

\paragraph{Triggering Mechanisms}
Condensers employ three paradigms: \emph{proactive} (\textit{RecentEvents, Amortized, LLM-based}) activate when event count exceeds \texttt{max\_size}; \emph{per-turn} (\textit{ObservationMasking}) applies transformations every iteration regardless of context size; \emph{reactive} (\textit{ConversationWindow}) activates only on explicit context overflow errors.
Each task runs with \texttt{max\_iterations=50} and \texttt{temperature=0.0} for deterministic outputs.

\section{Empirical Analysis}
\label{sec:evaluation}

We present experimental results evaluating 8 memory condensation strategies on DiscoveryBench, addressing the research questions posed in Section~\ref{sec:introduction}.




\begin{figure*}
  \centering
  \includegraphics[width=0.85\textwidth]{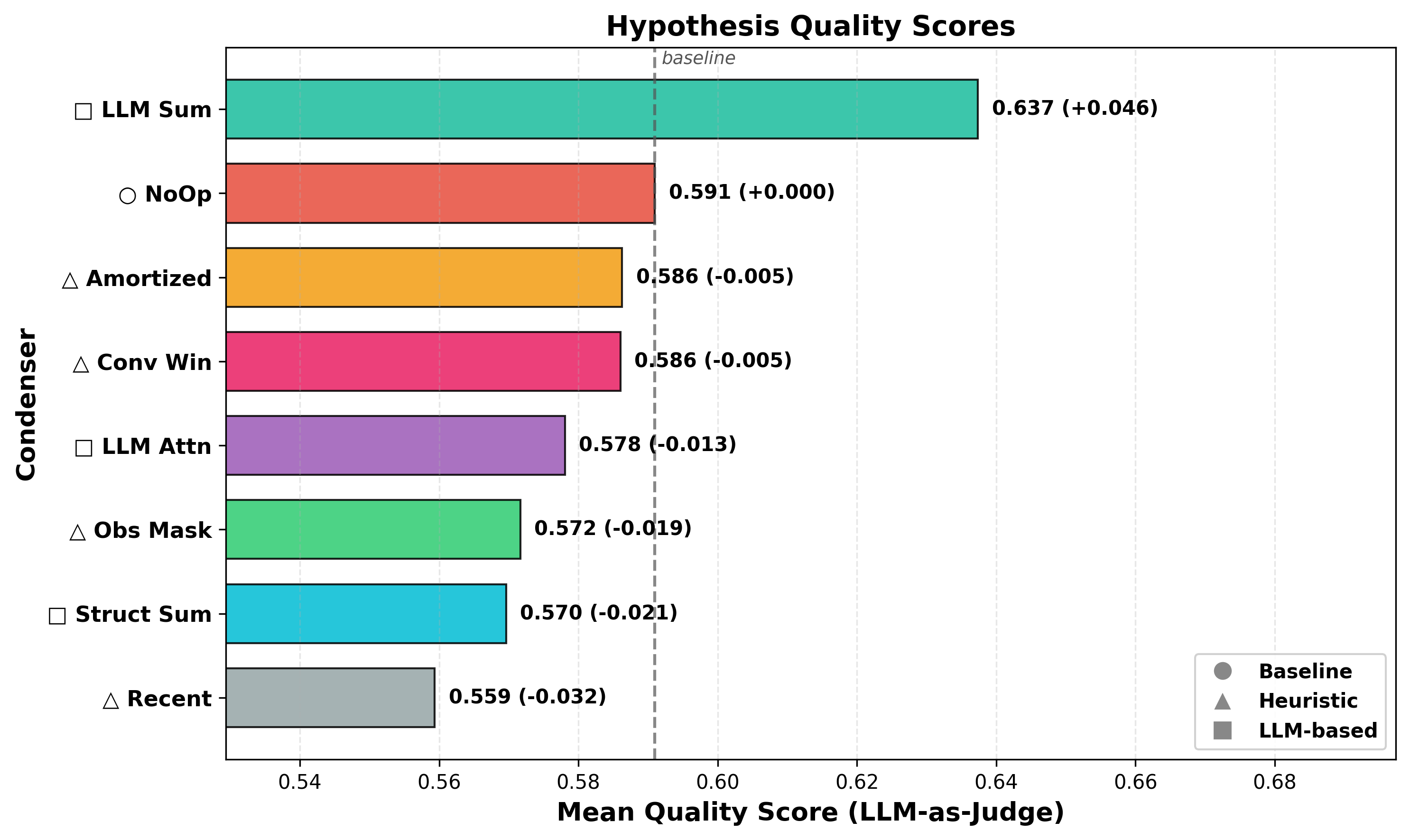}
  \caption{Hypothesis quality scores (LLM-as-Judge)}
  \label{fig:quality-scores}
\end{figure*}

\begin{table*}
  \centering
  \setlength{\tabcolsep}{3pt}
  \begin{tabular}{@{}lrrrrrr@{}}
    \toprule
    \textbf{Condenser} &
    \textbf{Average Tokens} &
    \textbf{Savings} &
    \textbf{Average Events} &
    \textbf{Faster} &
    \textbf{Same} &
    \textbf{Slower} \\
    \midrule
    ObservationMasking & 68,718 & \textbf{8.6\%}  & 9.6 & 32\% & 27\% & 41\% \\
    ConversationWindow & 74,243 & 1.2\%          & 9.8 & 30\% & 27\% & 43\% \\
    Amortized          & 77,824 & $-3.6\%$       & 12.2 & 18\% & 22\% & 60\% \\
    RecentEvents       & 85,971 & $-14.4\%$      & 10.0 & 35\% & 22\% & 43\% \\
    LLMSummarizing     & 92,880 & $-23.6\%$      & 13.3 & 33\% & 15\% & 52\% \\
    StructuredSummary  & 119,505 & $-59.0\%$     & 15.9 & 18\% & 12\% & 70\% \\
    LLMAttention       & 145,738 & $-93.9\%$     & 17.6 & 20\% & 13\% & 67\% \\
    NoOp (Baseline)    & 75,147 & ---            & 9.0 & --- & --- & --- \\
    \bottomrule
  \end{tabular}
  \caption{DiscoveryBench results (60 tasks × 8 condensers). Average total tokens and event counts per instance, with savings relative to NoOp baseline.}
  \label{tab:main-results}
\end{table*}

\subsection{RQ1: How does memory condensation impact task completion rates, token savings and solution quality?}
\label{sec:accuracy}

An important question is whether condensation degrades task performance. We evaluate this through task completion metrics and hypothesis quality assessment.

 \paragraph{Task Completion}                            All heuristic condensers maintain 100\% task completion; LLM-based condensers achieve 93--98\% completion. 
 However, we noticed that most methods increase total token usage relative to the \textit{NoOp} baseline. 
 Only \textit{ObservationMasking} (+8.6\%) and \textit{ConversationWindow} (+1.2\%) achieve net savings, while all others add 4--94\% more tokens due to condensation overhead and extra agent turns (see Table~\ref{tab:main-results}). 
 Across methods, 41--70\% of tasks require more events, with only 18--35\% completing faster. LLM-based condensers are especially inefficient: \textit{StructuredSummary} increases usage by 59\% (70\% tasks slower) and \textit{LLMAttention} by 94\% (67\% slower). These slowdowns stem from agents losing necessary context and retrying failed approaches, compounded by the added LLM inference overhead.

\begin{figure*}
\centering                                              
\includegraphics[width=\textwidth]{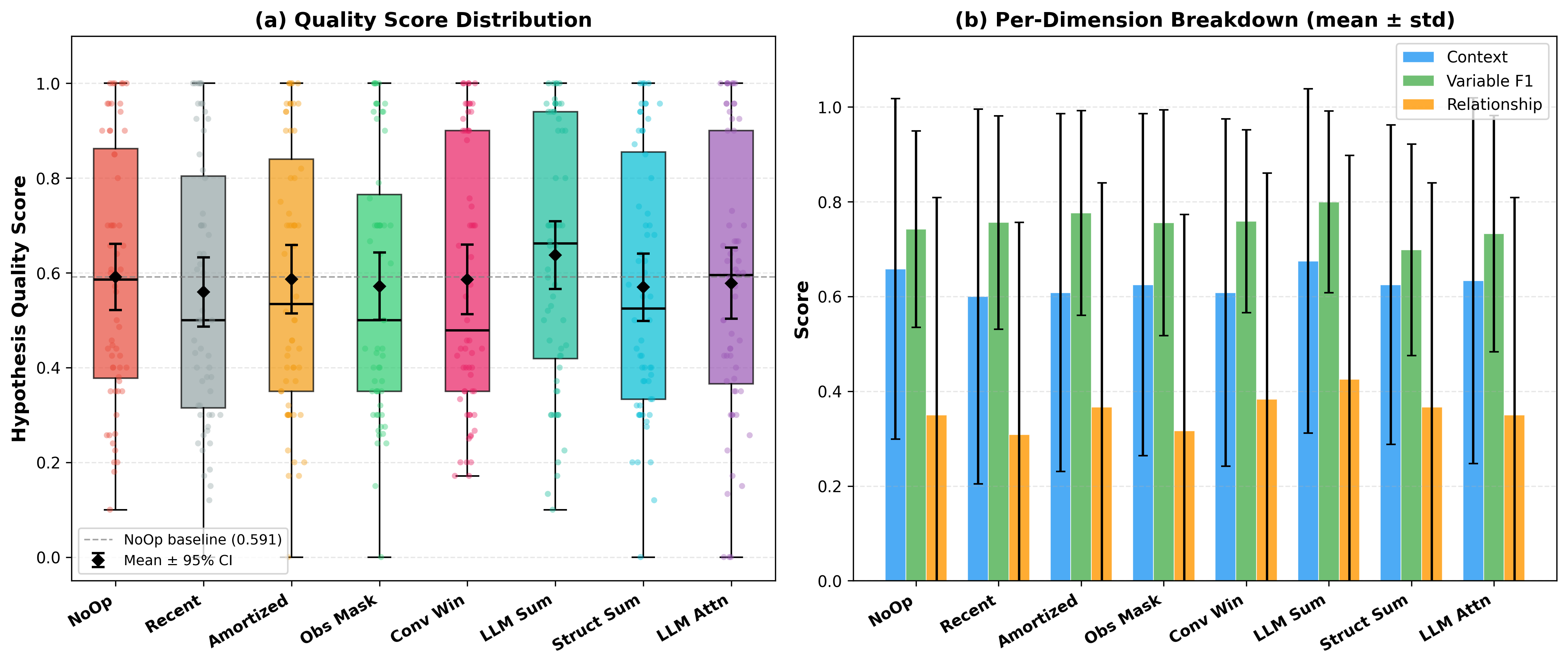}                        \caption{Hypothesis quality analysis. (a) Box plots with individual data points and mean $\pm$ 95\% CI (black         
diamonds). The dashed line marks the NoOp baseline (0.591) - All confidence intervals overlap (b) Per-dimension breakdown (mean $\pm$ std).            
Relationship scoring shows the highest variance across all condensers, while Variable F1 remains the most stable.}    
\label{fig:quality-detail}                              \end{figure*}

\paragraph{Hypothesis Quality}    
Hypothesis quality remains relatively comparable across condensers, ranging from 0.559 (\textit{RecentEvents}) to 0.637 (\textit{LLMSummarizing}), compared to the \textit{NoOp} baseline of 0.591 (Figure~\ref{fig:quality-scores}). As shown in Figure~\ref{fig:quality-detail}a, all confidence intervals overlap with the baseline, and paired Wilcoxon signed-rank tests confirm no significant differences (all $p > 0.05$, Bonferroni-corrected). \textit{LLMSummarizing} achieves the highest scores across all three dimensions (Context: 0.675 vs.\ 0.658 (baseline), Variable F1: 0.800 vs.\ 0.742 (baseline), Relationship: 0.425 vs.\ 0.350), indicating that summarization helps agents retain task-relevant information. \textit{RecentEvents} performs worst due to discarding older reasoning steps important for multi-step discovery. No condenser produces a statistically significant change in hypothesis quality. This shows memory condensation minimally affects hypothesis quality, and that cost, not quality - largely differentiates strategies.


\begin{figure}[!t]
\centering
\includegraphics[width=0.75\textwidth]{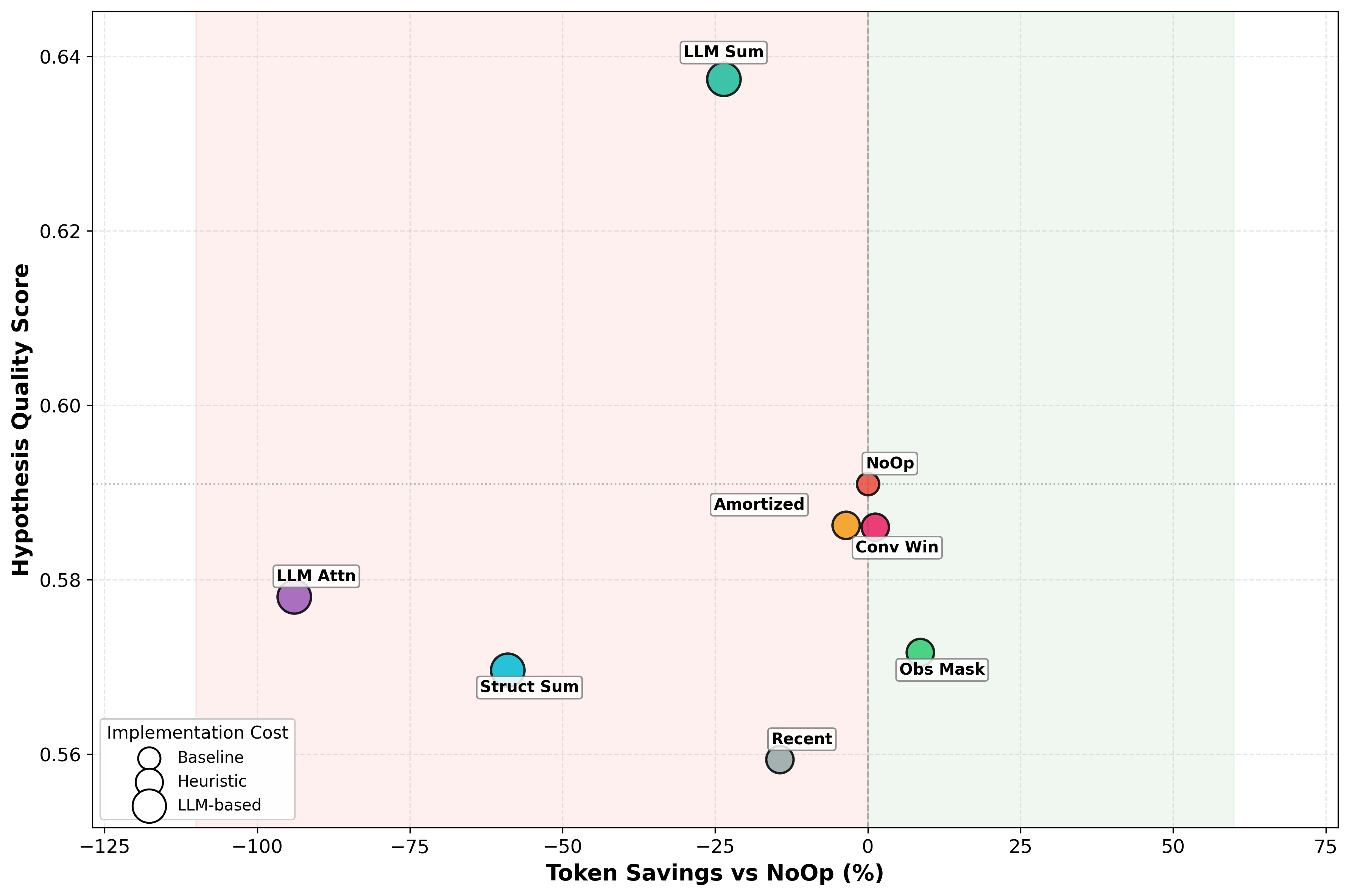}
\caption{Token Savings vs Hypothesis Quality.}
\label{fig:cost-benefit}
\end{figure}

\begin{figure}[!t]
\centering
\includegraphics[width=0.8\textwidth]{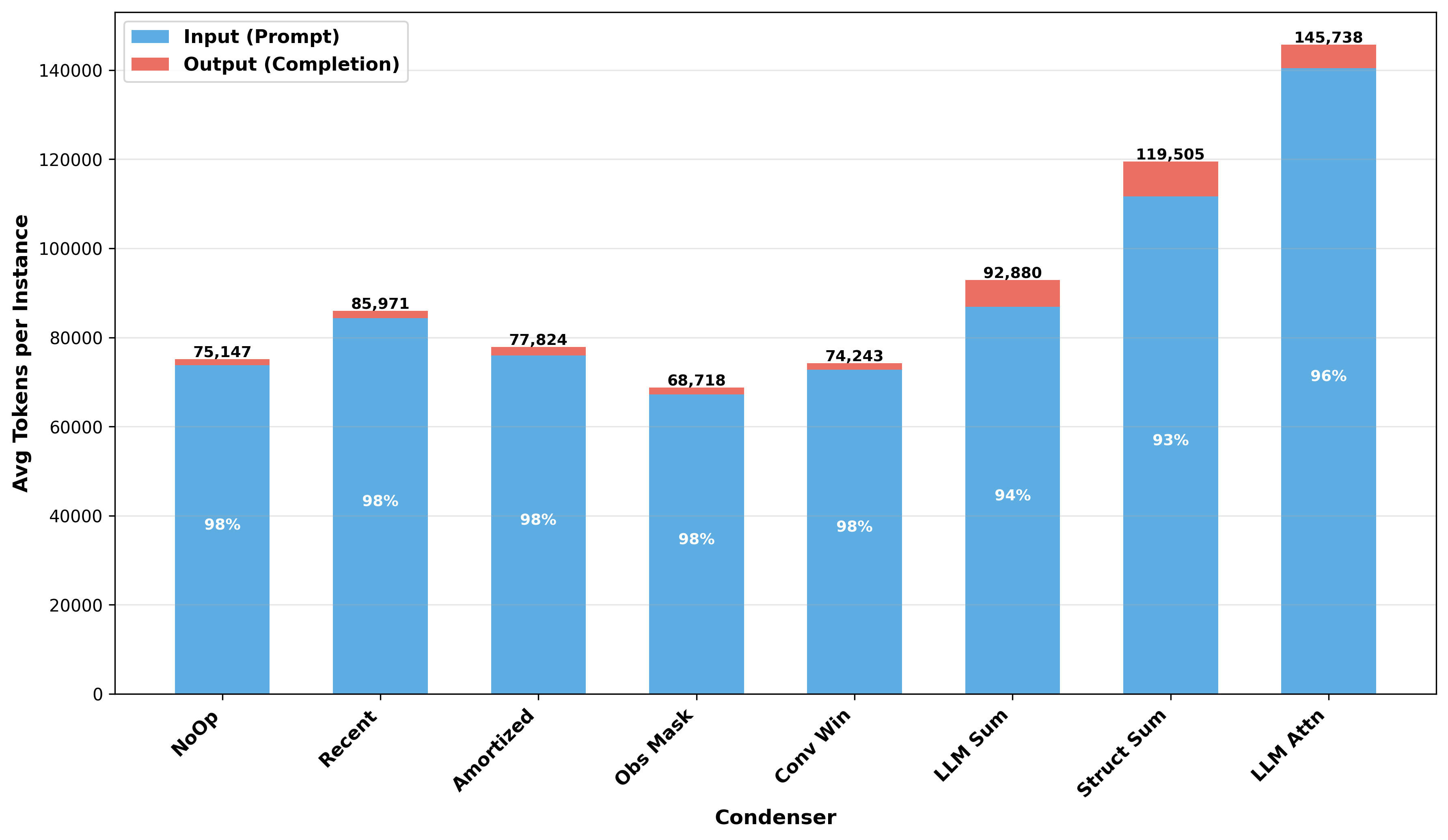}
\caption{Input vs.\ output token breakdown. Input tokens dominate at 93--98\% across all condensers. Cost savings come from total reduction, not ratio optimization.}
\label{fig:io-tokens}
\end{figure}

\paragraph{Token Savings vs. Hypothesis Quality}
Figure~\ref{fig:cost-benefit} reports the token savings and hypothesis quality across different memory condensation techniques after 60 runs each.
\textit{ObservationMasking} is the most efficient method in saving tokens with near‑baseline quality.
While LLM‑based methods do not achieve any token savings over the baseline, the \textit{StructuredSummary} method comes closest to matching baseline quality, whereas \textit{LLMSummarizing} most closely matches the baseline’s token savings.
In the context of \textit{LLMSummarizing}, we believe that developing effective token‑saving strategies will be useful to fully capitalize on its potential performance advantages.

Figure~\ref{fig:io-tokens} further shows that input tokens dominate total cost (94--98\%) for all condensers.
LLM-based strategies produce a slightly higher share of output tokens due to condensation steps, but input tokens remain the primary cost driver, highlighting that reducing total token count matters more than optimizing the input/output ratio.



\textbf{Key finding:} All heuristic condensers maintain 100\% task completion; LLM-based condensers achieve 93--98\% completion and have no statistically significant effect on hypothesis quality (Wilcoxon $p > 0.05$, Bonferroni-corrected). Scores range from 0.559
(\textit{RecentEvents}) to 0.637 (\textit{LLMSummarizing}) vs.\ 0.591 baseline, with all confidence intervals overlapping (Figure~\ref{fig:quality-detail}a). \textit{LLMSummarizing} best preserves reasoning chains, leading on all three dimensions - especially Relationship (0.425 vs.\ 0.350) - at 24\% higher token cost. \textit{ObservationMasking} offers the best efficiency - quality balance (8.6\% savings, 0.572 quality).

\FloatBarrier
\subsection{RQ2: Do different scientific domains benefit from different condensation approaches?}
\label{sec:domains}


\begin{figure*}
\centering
\begin{subfigure}[t]{0.48\textwidth}
\centering
\includegraphics[width=\textwidth]{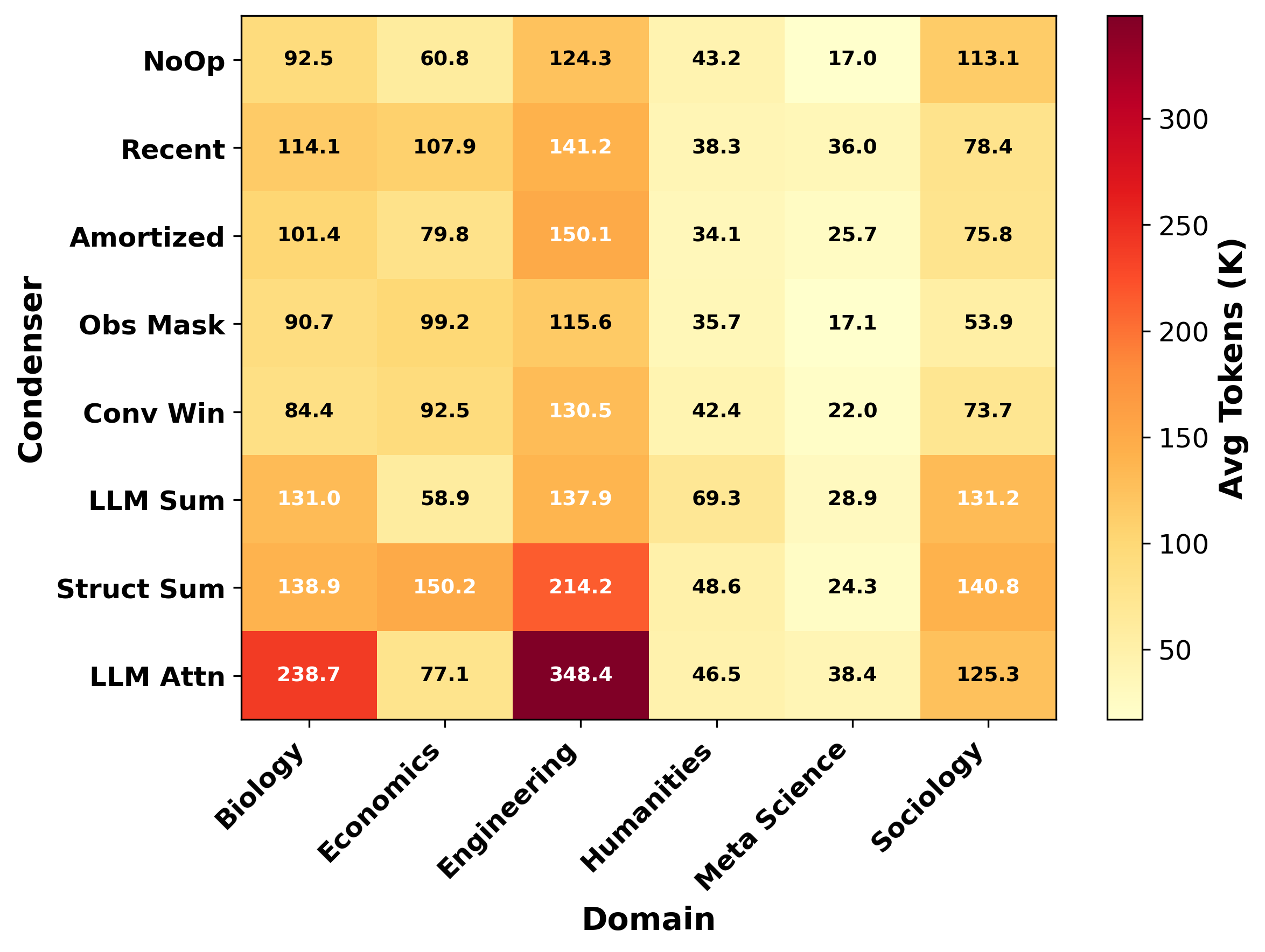}
\caption{Average tokens (K) by domain and condenser. Lighter colors indicate lower token usage.}
\label{fig:domain-heatmap}
\end{subfigure}%
\hfill
\begin{subfigure}[t]{0.48\textwidth}
\centering
\includegraphics[width=\textwidth]{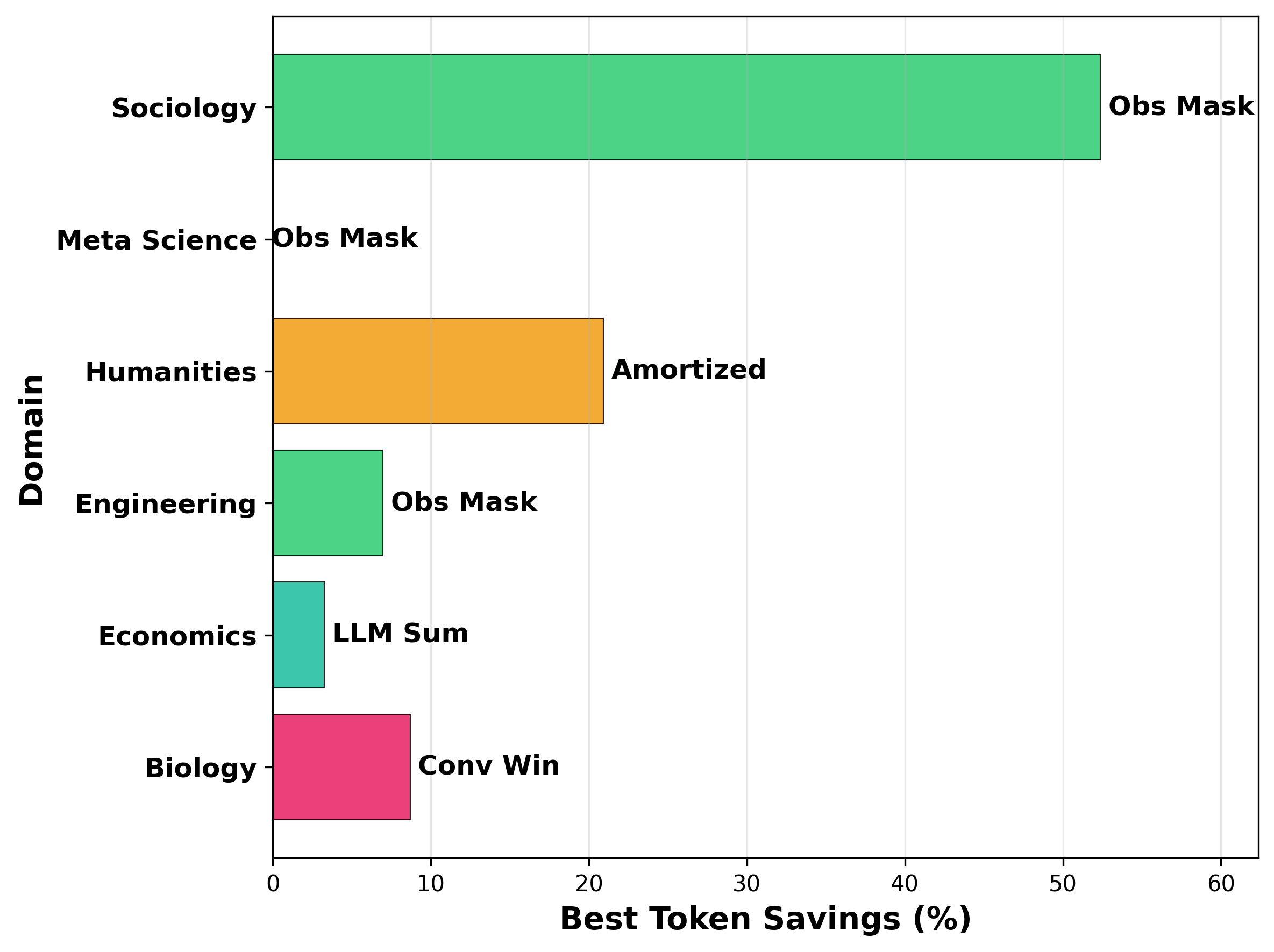}
\caption{Best condenser per domain with maximum savings achieved.}
\label{fig:domain-best}
\end{subfigure}
\caption{Domain-specific condenser performance analysis.}
\label{fig:domain-analysis}
\end{figure*}

\textbf{Key finding: No single condenser wins across domains.}
As shown in Figure~\ref{fig:domain-best}, Sociology benefits most from condensation (52\% savings with \textit{ObservationMasking}), as its verbose regression outputs contain large execution traces with sparse hypothesis-relevant content.
MetaScience resists condensation entirely, its compact, already-aggregated outputs leave little noise to remove.
Engineering also favors \textit{ObservationMasking} (7\%), whereas Economics is hardest to compress, with only \textit{LLMSummarizing} providing modest savings (3.3\%).
LLM-based methods show negative savings in Biology ($-158\%$) and Engineering ($-180\%$), where over-compression of multi-step statistical reasoning causes context loss and retries that compound condensation overhead.

\subsection{RQ3: How do token costs scale with task length across condensation strategies?}
\label{sec:growth}


\begin{figure*}[!t]
\centering

\begin{subfigure}[t]{\textwidth}
\centering
\includegraphics[width=0.60\textwidth]{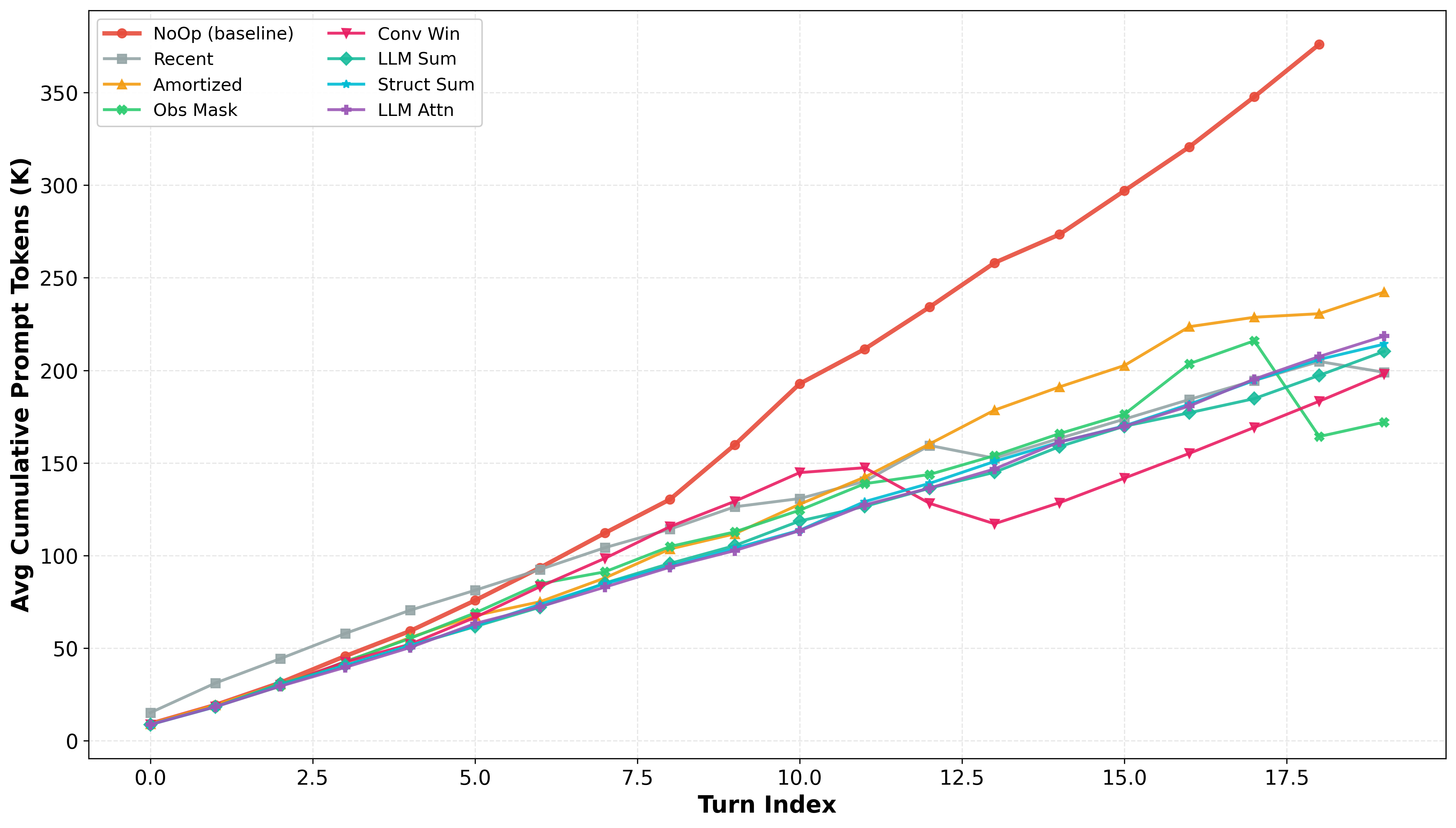}
\caption{Average cumulative prompt tokens per LLM turn, averaged across all 60 instances. NoOp (red) exhibits quadratic growth reaching ${\sim}$375\,K by turn~19, while all condensers maintain sub-linear token accumulation, with the gap widening progressively at later turns.}
\label{fig:token-accumulation-a}
\end{subfigure}
\vspace{0.5em}
\begin{subfigure}[t]{\textwidth}
\centering
\includegraphics[width=0.60\textwidth]{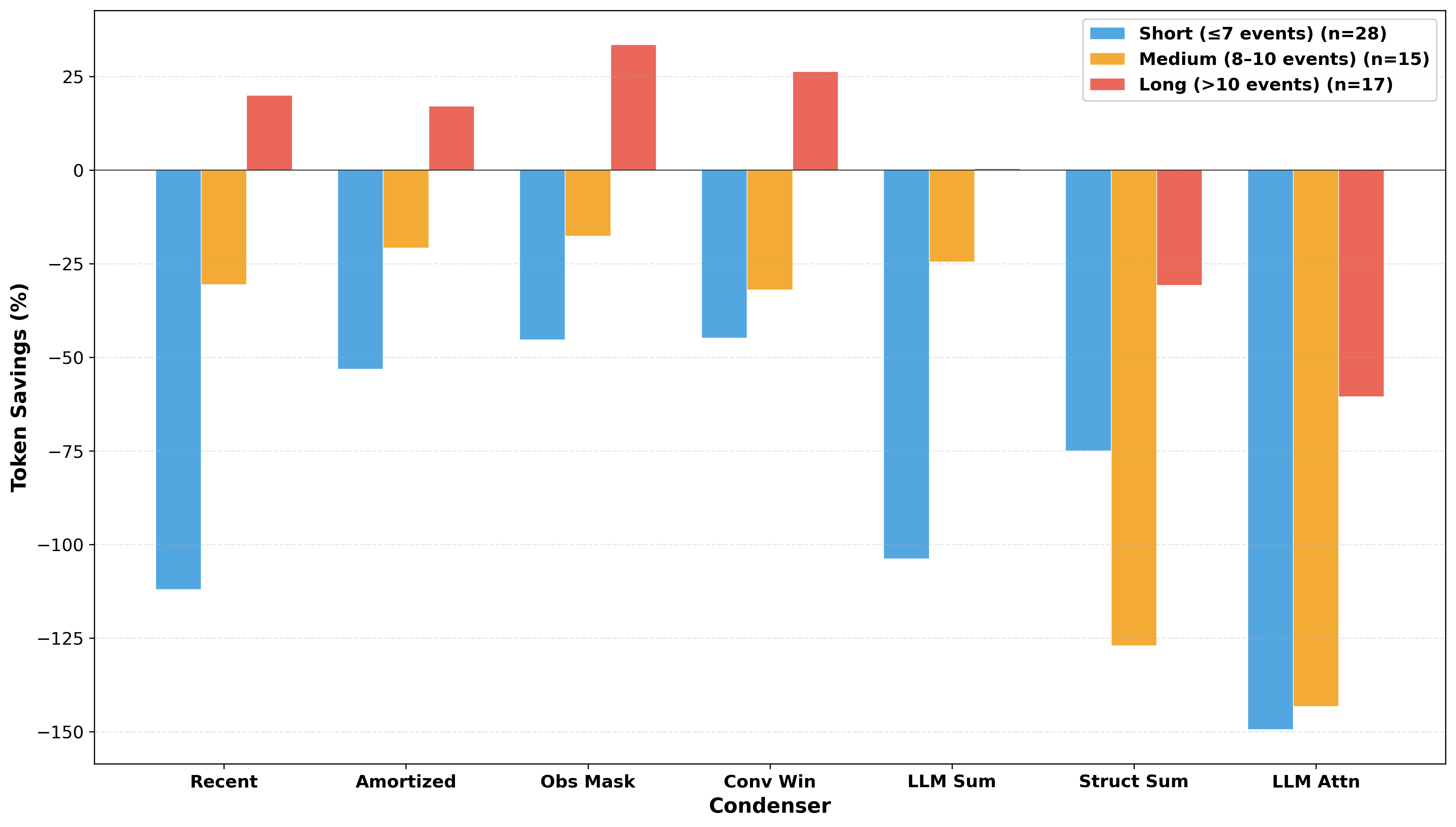}
\caption{Token savings (\%) by task length. Tasks are grouped as short ($\leq$7 events), medium (8--10 events), and long ($>10$ events) based on NoOp agent-event counts. Positive values indicate savings; negative values indicate overhead.}
\label{fig:token-accumulation-b}
\end{subfigure}
\caption{Token growth patterns demonstrating the efficiency benefits of condensation for long-running tasks.}
\label{fig:token-accumulation}
\end{figure*}

\textbf{Key finding: Memory condensation benefits grow with task length, but only heuristic methods record net
token savings---and only on long tasks.}
Figure~\ref{fig:token-accumulation}a shows the average cumulative prompt tokens per turn.
Without condensation (\textit{NoOp}),
cumulative cost grows steeply from $\sim$9\,K tokens at turn~0 to $\sim$93\,K by turn~6 and $\sim$376\,K by
turn~18.
All condensers track close to \textit{NoOp} through the first 4--5 turns, after which the lines diverge: \textit{ObservationMasking} and \textit{ConversationWindow} accumate tokens in a
slower rate, reaching roughly half the \textit{NoOp} total by turn~15.
Figure~\ref{fig:token-accumulation}b breaks this pattern down by task duration (Short $\leq 7$ events, $n=28$;
Medium 8--10, $n=15$; Long $>10$, $n=17$).
Short tasks incur overhead for every condenser ($-45\%$ to $-149\%$)
because condensation costs cannot be amortized over few turns.
Medium tasks still show net increases across the
board.
Only long tasks ($>10$ events) record true savings where \textit{ObservationMasking} $+33.5\%$,
\textit{ConversationWindow} $+26.3\%$, \textit{RecentEvents} $+19.9\%$, and \textit{Amortized} $+17.0\%$ have token savings.
LLM-based condensers remain net-negative even on long tasks (\textit{StructuredSummary} $-30.8\%$,
\textit{LLMAttention} $-60.5\%$), with only \textit{LLMSummarizing} roughly breaking even ($+0.4\%$).

\section{Conclusion}
\label{sec:conclusion}

We evaluated 8 memory condensation strategies across 480 DiscoveryBench instances in 6 scientific domains. No condenser produces a statistically significant change in hypothesis quality, while LLM-based condensers increase total token cost by 24--94\%. \textit{ObservationMasking} achieves the largest net savings (8.6\%) without degrading quality. All heuristic condensers maintain 100\% task completion, while LLM-based condensers achieve 93--98\%. Quality varies only modestly (0.559--0.637), indicating that cost, not quality - differentiates strategies. Domain context matters substantially: \textit{ObservationMasking} saves 52\% on verbose sociology tasks but provides no benefit on compact meta-science tasks. Only long tasks ($>$10 events) yield positive savings for heuristic condensers (17--34\%), while short tasks see increased costs across all strategies. 
These findings motivate adaptive meta-condensers that select or blend strategies at runtime based on domain, task length, and output verbosity.

\section{Limitations}
\label{sec:limitations}

Our study has three main limitations. 
First, all evaluations use a single backbone model (GPT-4o) with GPT-5-mini for LLM-based condensation; condensation effectiveness may differ for models with smaller context windows or weaker instruction-following capabilities.
Second, DiscoveryBench tasks range from 5--22 events; real-world scientific workflows involving hundreds of iterations may amplify or shift the patterns we observe, particularly the task-length threshold at which condensation becomes cost-effective.
Third, we treat condenser selection as a static, per-run choice; in practice, agents could benefit from switching strategies mid-task as context demands evolve, a direction our findings on task-length and domain dependence directly motivate.


\section{Acknowledgments}
This work was supported by the U.S. Department of Energy, Advanced Scientific Computing Research
program and Pacific Northwest National Laboratory (PNNL), which is operated by Battelle Memorial Institute for the U.S. Department of Energy under Contract DE-AC05–76RLO1830. This research used resources of the National Energy Research Scientific Computing Center (NERSC), a Department of Energy User Facility using NERSC award ASCR-ERCAP0038273.
This manuscript is identified as PNNL-SA-222584. 
\bibliographystyle{colm2026_conference}
\bibliography{memory_references}


\end{document}